\documentclass[10pt,twocolumn,letterpaper]{article}

\usepackage{cvpr}
\usepackage{times}
\usepackage{epsfig}
\usepackage{graphicx}
\usepackage{amsmath}
\usepackage{amssymb}

\usepackage{amsmath,amssymb}
\usepackage{tabularx}
\usepackage{color}

\usepackage{url}
\usepackage{cite}

\usepackage{pdfpages}

\usepackage[pagebackref=true,breaklinks=true,letterpaper=true,colorlinks,bookmarks=false]{hyperref}

\cvprfinalcopy 

\def\cvprPaperID{1835} 

\newcommand\blfootnote[1]{%
  \begingroup
  \renewcommand\thefootnote{}\footnote{#1}%
  \addtocounter{footnote}{-1}%
  \endgroup
}
\begin{document}

\title{NDDR-CNN: Layerwise Feature Fusing in Multi-Task CNNs \\ by Neural Discriminative Dimensionality Reduction}

\author{Yuan Gao$^{1*}$ \quad Jiayi Ma$^2$ \quad Mingbo Zhao$^3$ \quad Wei Liu$^{1*}$ \quad Alan L. Yuille$^4$\\
$^1$ Tencent AI Lab ~
$^2$ Wuhan University ~
$^3$ City University of Hong Kong ~
$^4$ Johns Hopkins University \\
{\tt\footnotesize \{ethan.y.gao, jyma2010, mbzhao4\}@gmail.com, wl2223@columbia.edu, alan.yuille@jhu.edu}
}

\maketitle

\begin{abstract}
   In this paper, we propose a novel Convolutional Neural Network (CNN) structure for general-purpose multi-task learning (MTL), which enables automatic feature fusing at every layer from different tasks. This is in contrast with the most widely used MTL CNN structures which empirically or heuristically share features on some specific layers (\eg, share all the features except the last convolutional layer). The proposed layerwise feature fusing scheme is formulated by combining existing CNN components in a novel way, with clear mathematical interpretability as discriminative dimensionality reduction, which is referred to as Neural Discriminative Dimensionality Reduction (NDDR). Specifically, we first concatenate features with the same spatial resolution from different tasks according to their channel dimension. Then, we show that the discriminative dimensionality reduction can be fulfilled by 1 $\times$ 1 Convolution, Batch Normalization, and Weight Decay in one CNN. The use of existing CNN components ensures the end-to-end training and the extensibility of the proposed NDDR layer to various state-of-the-art CNN architectures in a ``plug-and-play'' manner. The detailed ablation analysis shows that the proposed NDDR layer is easy to train and also robust to different hyperparameters. Experiments on different task sets with various base network architectures demonstrate the promising performance and desirable generalizability of our proposed method. The code of our paper is available at \url{https://github.com/ethanygao/NDDR-CNN}.
\end{abstract}

\section{Introduction}

Deep\blfootnote{* indicates corresponding authors.} convolutional neural networks (CNNs) have greatly pushed the previous limits of various computer vision tasks since the seminal work \cite{krizhevsky2012imagenet} in 2012. CNN models can naturally integrate hierarchical features and classifiers, which can be trained in an end-to-end manner. Benefiting from that, significant improvements have been witnessed in fundamental computer vision tasks, such as image classification \cite{krizhevsky2012imagenet,simonyan2014very,he2016deep,huang2016densely,hu2017}, object detection \cite{girshick2014rich,girshick2015fast,ren2015faster,he2017mask,redmon2016yolo9000,liu2016ssd,tdm_arxiv16,adversarial_detection_cvpr17,primingfeedback_eccv16,ohem_cvpr16,tang2017multiple,tang2018pcl}, semantic segmentation \cite{chen2016deeplab,long2015fully,ronneberger2015u,badrinarayanan2015segnet,papandreou2015weakly,chen2016attention,chen2016semantic,xia2015zoom,chen2014semantic}, \etc. 

One of the main factors that can further boost the CNN performance is multi-task learning (MTL), which is engaged in learning multiple related tasks simultaneously. This is because related tasks can benefit from each other by jointly learning certain shared, or more precisely, mutually related representations \cite{he2017mask,kokkinos2016ubernet}. The multiple supervision signals originating from different tasks in MTL can be viewed as implicit data augmentation (on labels) or additional regularization (among different tasks) \cite{ruder2017overview}. This enables to learn mutually related representations that work well for multiple tasks, thus avoiding overfitting and leading to better generalizability.

Most commonly, the CNN structure for MTL is heuristically determined by sharing all convolutional layers, and splitting at fully-connected layers for task-specific losses. However, as different layers learn low-, mid-, and high-level features \cite{zeiler2014visualizing}, a natural question arises: \emph{Why would we assume that the low- and mid-level features for different tasks in MTL should be identical, especially when the tasks are loosely related? If not, is it optimal to share the features until the last convolutional layer?}

The study in Misra \etal \cite{misra2016cross} reveals that sharing/splitting at different layers gives different performances. Especially, improper features sharing at some layers may degrade the performance of some, or even all, tasks. In addition, the deep nature of CNNs makes it infeasible to exhaustively test all the possible structures to find the optimal sharing/splitting scheme. In order to tackle this issue, Misra \etal used trainable \emph{scalars} to weighted-sum the features from different tasks at multiple CNN levels and achieved state-of-the-art performance \cite{misra2016cross}.

We consider this problem in another way, \ie, by leveraging all the hierarchical features from different tasks. This is because that the CNN layers trained by different tasks can be treated as different feature descriptors, therefore the features learned from them can be treated as different representations/views of input data. \emph{We hypothesize that these features, obtained from multiple feature descriptors (\ie, different CNN levels from multiple tasks), contain additional discriminative information of input data, which should be exploited in MTL towards better performance}. 

Specifically, starting with $K$ single-task networks (from $K$ tasks), a direct attempt to take advantage of hierarchical features from all the tasks is that: we may concatenate all the task-specific features with the same spatial resolution \emph{from different tasks} according to the feature \emph{channel} dimension. After that, we expect the CNN to learn a discriminative feature embedding for each task, by receiving these concatenated features as inputs. However, most existing CNNs have carefully designed structures, which only receive features (tensors) with a fixed number of feature channels. By concatenating features, we substantially enlarge the number of channels as $K$ times if we have $K$ tasks. This makes it impossible to feed these concatenated features to the following layers of the CNN.

This property of the CNN motivates us to conduct \emph{discriminative dimensionality reduction} on the concatenated features. Its purpose is to learn a discriminative feature embedding, and to reduce the feature dimension such that it can satisfy the input channel requirement of the following layers. Feature transformation is one of the most important approaches to tackle the discriminative dimension reduction problem. It aims to learn a projection matrix that projects the original high-dimensional features into a low-dimensional representation, while keeping as much discriminative information as possible. 

In this paper, we show that, from the perspective of feature transformation, discriminative dimensionality reduction is closely related to some common operations of modern CNNs. Specifically, the transformation in discriminative dimensionality reduction is in fact equivalent to the \emph{$1 \times 1$ convolution}. In addition, the constraints on the norm of the transformation weights (\ie, the weights of the  $1 \times 1$ convolutional layer) and input feature vectors can be represented by \emph{weight decay} and \emph{batch normalization} \cite{ioffe2015batch}, respectively. We refer to the combination of these operations as \emph{Neural Discriminative Dimensionality Reduction} (NDDR).
Therefore, we are able to link the original single-task networks from different tasks by the NDDR layers. Desirably, the proposed network structure can be trained end-to-end in the CNN without any extraordinary operations.

It is worth noting that this paper focuses on a general structure for general-purpose MTL. The proposed NDDR layer combines existing CNN components in a novel way, which possesses clear mathematical interpretability as discriminative dimensionality reduction. Moreover, the use of the existing CNN components is desirable to guarantee the extensibility of our method to various state-of-the-art CNN architectures, where the proposed NDDR layer can be used in a ``plug-and-play'' manner. The rest of this paper is organized as follows. First, we describe the NDDR layer and propose a novel NDDR-CNN as well as its variant NDDR-CNN-Shortcut for MTL in Sect. \ref{Sect:methods}. After that, we discuss the related works in Sect. \ref{Sect:related_works}, where we show that our method can generalize several state-of-the-art methods, which can be treated as our special cases. In Sect. \ref{Sect:AA}, the ablation analysis is performed, where the hyperparameters used in our network are suggested. Following that, the experiments are performed on different network structures and different task sets in Sect. \ref{Sect:exp}, demonstrating the promising performance and desirable generalizability of our proposed method. We make concluding remarks in Sect. \ref{Sect:conclusion}.

\section{Related Works \label{Sect:related_works}}
Various computer vision tasks benefit from MTL \cite{sermanet2013overfeat}, such as detection \cite{girshick2014rich,girshick2015fast,ren2015faster,he2017mask,tdm_arxiv16,adversarial_detection_cvpr17,primingfeedback_eccv16,ohem_cvpr16}, human pose and semantic segmentation \cite{xia2017joint}, surface normal prediction, depth prediction, semantic segmentation \cite{eigen2015predicting}, action recognition \cite{yang2017discriminative,yang2017latent}, \etc. Several human face related tasks, including face landmark detection, attributes detection (such as smile and glasses), gender classification, and face orientation, were studied in \cite{trottier2017multi,han2017heterogeneous,ranjan2017hyperface}. Yim \etal used face alignment and reconstruction as auxiliary tasks for face recognition \cite{yim2015rotating}. MTL on sequential data was also studied in \cite{liu2016deep}. Recently, Kokkinos proposed a UberNet which enables a great number of low-, mid-, and high-level vision tasks to be handled simultaneously \cite{kokkinos2016ubernet}.

CNN based MTL theory has also been greatly developed in recent years. Long and Wang proposed a deep relationship network to enable the feature sharing at the fully-connected layers \cite{long2015learning}. Starting with a thin network, a top-down layerwise widening method was proposed to automatically determine which layer to split \cite{lu2016fully}. Yang and Hospedales used tensor decomposition at initialization to share the MTL weights \cite{yang2016deep}. The weights to combine the task-specific losses were also studied, and a Bayesian approach was proposed to predict these weights \cite{kendall2017multi}. The cross-stitch network used \emph{trainable scalars} to fuse (\ie, weighted sum) the features at layers in the same level from different tasks \cite{misra2016cross}. Most recently, the sluice network predefines several subspaces on the features from each task and learns the weights to fuse the features across different subspaces \cite{ruder2017sluice}.

Our method is also related to discriminative dimensionality reduction. The goal of the discriminative dimensionality reduction techniques is to reduce the computational and storage costs, by learning a low-dimensional embedding that retains most of the discriminative information. Linear discriminant analysis (LDA) is one of the most popular conventional discriminative dimensionality reduction methods, which aims to seek the optimal projection matrix by maximizing the between-class variance and meanwhile minimizing the within-class variance \cite{Martinez01}. In addition, low-rank metric learning \cite{liu2015low} can also be viewed as a discriminative dimensionality reduction technique.

Introduced by network in network \cite{lin2013network}, $1 \times 1$ convolution has been widely used in many modern CNN architectures \cite{szegedy2016rethinking,he2016deep,huang2016densely,Lin_2017_CVPR}. For example, it was used in ResNet to reduce the number of weights to train, by producing a ``bottleneck unit'' \cite{he2016deep}. $1 \times 1$ convolution is also implemented in the feature pyramid network to fuse hierarchical features (in different CNN levels) on a single task \cite{Lin_2017_CVPR}. Note that we do NOT claim the $1 \times 1$ convolution as our novelty. Instead, we use $1 \times 1$ convolution together with batch normalization and weight decay in a novel way, which yields an NDDR layer. In other words, we formulate the multi-task feature fusing paradigm as a discriminative dimensionality reduction problem, and use the NDDR layer, which is composed of $1 \times 1$ convolution, batch normalization, and weight decay, to learn the feature embeddings from multiple tasks. The use of the existing CNN components ensures the extensibility of our method to various state-of-the-art CNN architectures in a ``plug-and-play'' manner.

\section{Methodology \label{Sect:methods}}
In this section, we propose a novel method to automatically learn the optimal structure for layerwise feature fusing in a multi-task CNN. Instead of the ``split-style'' multi-task CNN (\eg, split at the last convolutional layer for different task-specific losses), we consider the ``fuse-style'' network combining multiple single-task networks via discriminative dimensionality reduction. 

We first relate the discriminative dimensionality reduction to $1 \times 1$ convolution and propose the NDDR layer. Then, a novel multi-task network is proposed, namely NDDR-CNN, where the NDDR layer is leveraged to connect the original single-task networks. Moreover, a variant of NDDR-CNN is introduced, namely NDDR-CNN-Shortcut, which enables to directly route the gradients to the lower NDDR layers by shortcut connections. Finally, we give the implementation details of the proposed network. 

\subsection{NDDR Layer}
As discussed in previous sections, we aim to utilize the hierarchical features learned from different tasks. It is unlike the most widely used method which heuristically shares all the low-(and mid-) level features and splits the network at the last convolutional layer.


\begin{figure*}[t]
\centering
\includegraphics[width=0.9\linewidth]{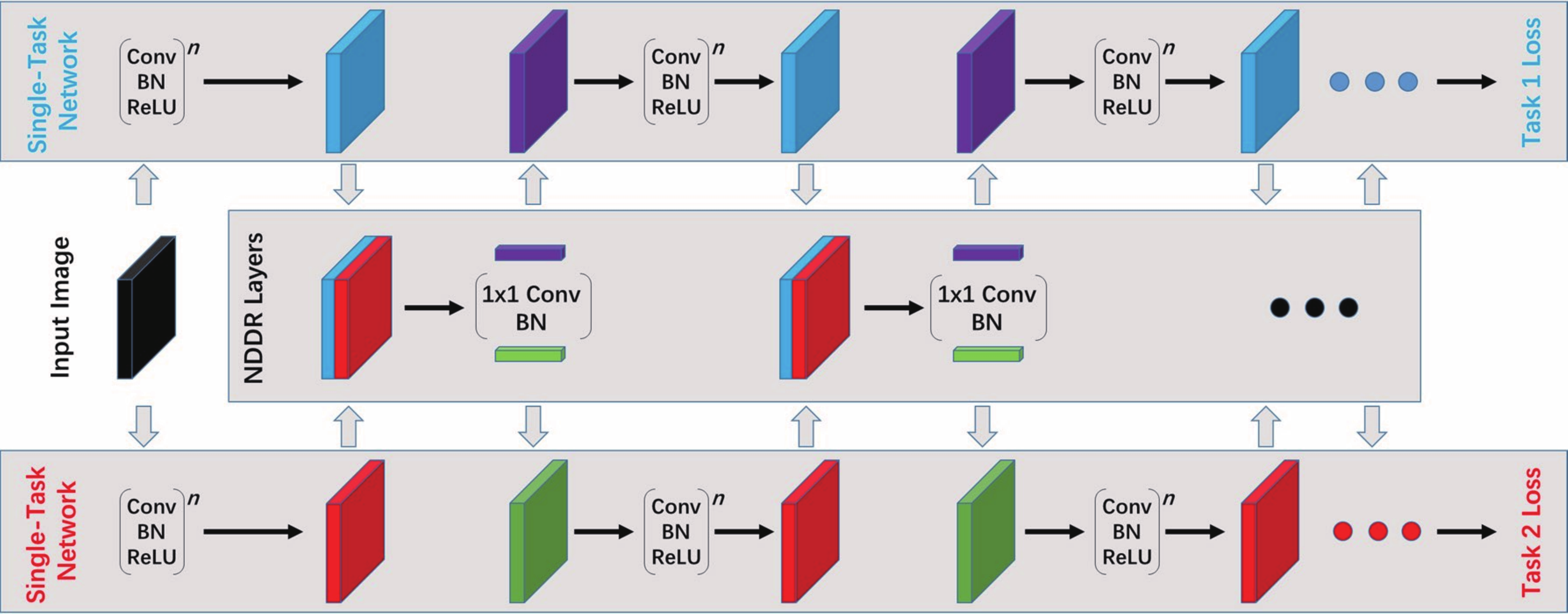}
\caption{The network structure of NDDR-CNN. In the NDDR layer, we concatenate the outputs of original single-task networks from multiple tasks (two tasks shown here), and use $1 \times 1$ convolution to perform discriminative dimensionality reduction. Therefore, the output of the NDDR layer retains the discriminative information from both the input features, and can be fed to the following layers of the single-task networks. 
The proposed NDDR layer can be leveraged to connect the original single-task networks of multiple levels for layerwise feature fusing (best view in color).}
\label{Fig:NDDR-CNN}
\end{figure*}

In order to do that, we first concatenate the task-specific features from different tasks according to the channel dimension. Then, we use a discriminative dimensionality reduction technique to reduce the feature channels such that the output features satisfy the channel dimension requirement of the next CNN layers. We refer to the new CNN layer with such operations as the Neural Discriminative Dimensionality Reduction (NDDR) layer.

Specifically, let $F^i_l \in \mathbb{R}^{N \times H \times W \times C}$ be the output features (arranged in a tensor) at an intermediate layer $l$ of task $i$. Regarding $K$ tasks, concatenating the features from them according to the channel dimension gives: 
\begin{equation}
  F_l = [F^1_l, ..., F^K_l] \in \mathbb{R}^{N \times H \times W \times KC}.
\end{equation}

Discriminative dimensionality reduction learns a transformation $W$ to reduce the dimensionality of the input features, while keeping most discriminative information:
\begin{equation}
  F_l^{i*} = F_l W^{i},
\end{equation}
where $W^{i} \in \mathbb{R}^{KC \times M}$ and $M < KC$ is the projection matrix to be learned for each task $i$. In our case, $M$ is equal to $C$ (\ie, $F_l^{i*} \in \mathbb{R}^{N \times H \times W \times C}$) in order to satisfy the channel size requirement of the following CNN layers.

Conventional discriminative dimensionality reduction methods learn the transformation $W$ with specific assumptions/objectives which make the features more separable. For example, Linear Discriminative Analysis (LDA) learns $W$ by minimizing the projected \emph{within-class} variation and meanwhile maximizing the projected \emph{between-class} variation \cite{Martinez01}. Intuitively, the objective function of the discriminative dimensionality reduction is related to the CNN loss, \ie, the features projected by discriminative dimensionality reduction are more separable, therefore giving a smaller CNN loss.

Motivated by this, we aim to learn the transformation $W$ in the CNN implicitly by back-propagation. The transformation $W \in \mathbb{R}^{KC \times C}$ can be represented precisely by a convolution operation with stride 1 and size ($C \times 1 \times 1 \times KC$), where these size dimensions represent filters, kernel height, kernel width, and channels, respectively. It is worth noting that the convolution with $1 \times 1$ kernel size and 1 stride enables to perform the computations only according to channels, rather than fusing the features at different spatial locations or changing the spatial sizes of the features.

In addition, discriminative dimensionality reduction methods also have constraints on the norms of the transformation $W$ (to avoid a trivial solution) and the input features $F_l$ (otherwise, the learned projections may project the features to some \emph{noise} directions). We borrow this idea to our NDDR layer for stable learning, which can be achieved by imposing \emph{batch normalization} on the input features and $\ell_2$ \emph{weight decay} on the $1 \times 1$ convolutional weights $W$, respectively.

In summary, a novel NDDR layer is proposed in this section. The NDDR layer can be constructed by: 1) concatenating the task-specific features with the same spatial resolution from different tasks according to the channel dimension, and 2) using \emph{$1 \times 1$ convolution} to learn a discriminative feature embedding for each task. We also use \emph{batch normalization} on the input features of the NDDR layer for stable learning. We train the NDDR layer by back-propagating the \emph{task-specific} losses and the $\ell_2$ \emph{weight decay} loss on the $1 \times 1$ convolutional weights $W$. \emph{Without any extraordinary operations, the network with our NDDR layer can be trained in an end-to-end fashion}.

\subsection{NDDR-CNN Network}
We insert the NDDR layers in multiple levels of the original single-task networks, to enable layerwise feature fusing/embedding for different tasks. We refer to the proposed network for MTL as the NDDR-CNN network. 

Figure \ref{Fig:NDDR-CNN} shows the NDDR-CNN network structure for two tasks. It can easily be extended to $K$-task problems. Let the number of channels for the single-task features be $D$. Then NDDR-CNN for $K$ tasks can be constructed by: 1) concatenating the features from $K$ tasks according to the \emph{channel} dimension, and 2) using $1 \times 1$ convolution with (filters $ \times 1 \times 1 \times $ channels) = ($C \times 1 \times 1 \times KC$) to conduct dimensionality reduction, where $C$ is the channel dimension size of the output features from each task.

Note that the elements of the NDDR layer are common CNN operations, which ensures that the proposed NDDR layer can be extended to various state-of-the-art CNN architectures in a ``plug-and-play'' manner.

\subsection{NDDR-CNN Network with Shortcuts}
In order to avoid gradient vanishing at lower NDDR layers, we propose a new network that enables to pass gradients directly from the last convolutional layer to the lower ones via \emph{shortcut connections}, namely NDDR-CNN-Shortcut. 

\begin{figure}[t]
\centering
\includegraphics[width=\linewidth]{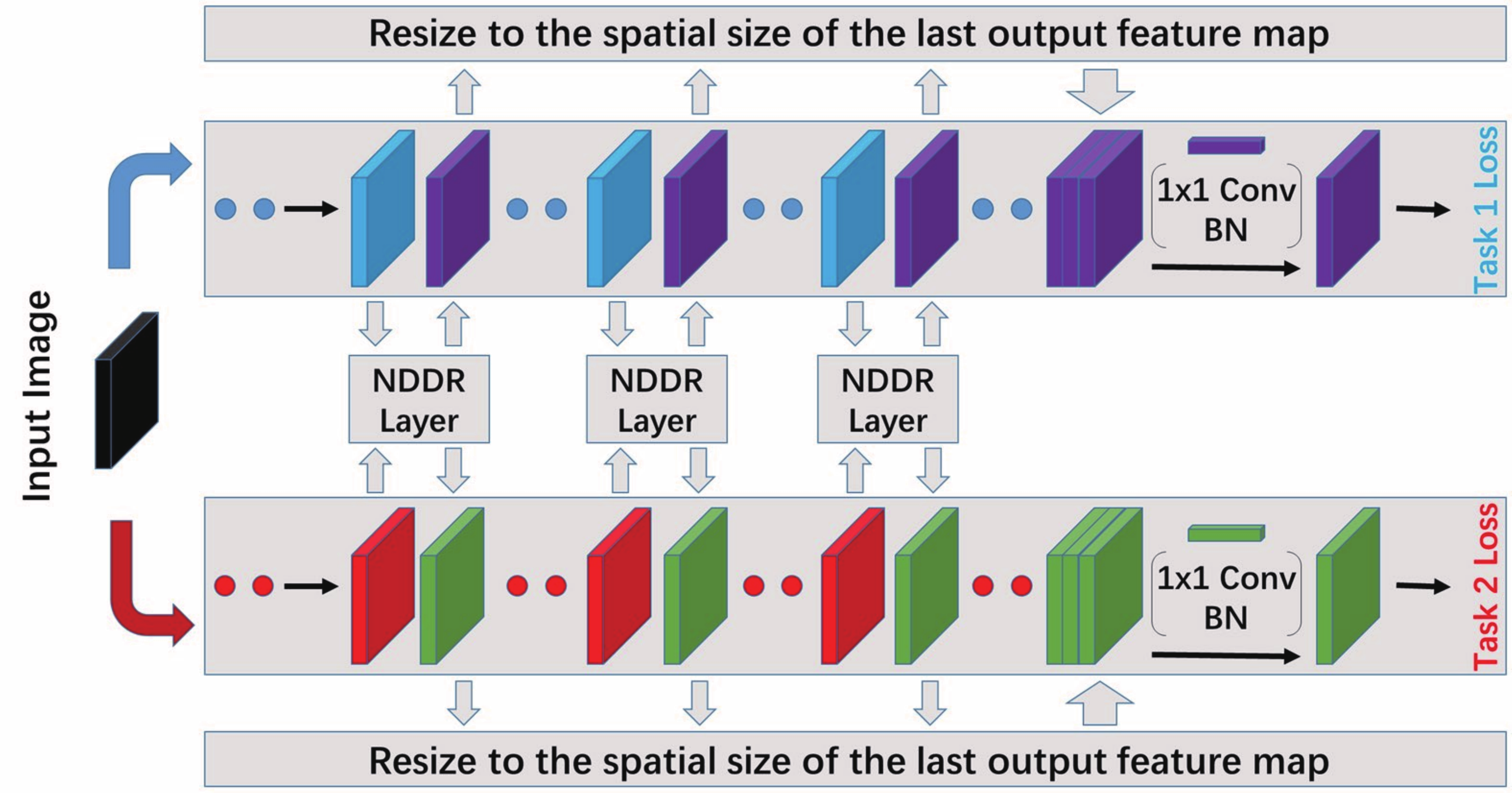}
\caption{The NDDR-CNN-Shortcut network. In NDDR-CNN-Shortcut, we use shortcut connections to enable gradients to directly route to the lower NDDR layers. This is done by resizing the lower NDDR output to the spatial size of the last NDDR output, then concatenating the resized features of the same task according to the channel dimension, and finally using $1 \times 1$ convolution to do dimensionality reduction (best view in color).}
\label{Fig:NDDR-CNN-Short}
\end{figure}

Specifically, the output of each NDDR layer is resized to the spatial sizes of the last convolutional output. Then we concatenate all the resized feature maps of the same task from different layers together according to the channel dimension. Finally, in order to fit the input size of the following fully-convolutional/connected layers, we further use $1 \times 1$ convolution to learn more compact feature tensors (\eg, in the VGG network, we reduce the channel dimension of concatenated features to 512). An illustration of the NDDR-CNN-Shortcut network is shown in Fig. \ref{Fig:NDDR-CNN-Short}.

\subsection{Relationship to State-of-the-art Methods}
Our method is closely related to the cross-stitch network \cite{misra2016cross}. In order to seek the optimal network structure for MTL, the cross-stitch network \cite{misra2016cross} uses \emph{trainable scalars} to scale the features at layers in the same level from different tasks, and then adds them together as new features. Our work is related to the cross-stitch network but has three major differences: 1) We have different motivations, \ie, our work is motivated by learning discriminative low-dimensional embeddings on the concatenated features from multiple tasks. 2) Our method can generalize the cross-stitch network \emph{by fixing the off-diagonal elements of the projection matrix to 0, and only updating the diagonal elements with the same value} (\ie, update the projection matrix by $\alpha$ and $\beta$ in Eq. \eqref{init}). 3) We further propose an NDDR-CNN-Shortcut model, which further uses hierarchical features from different CNN levels for better training and convergence. Similarly, our network also takes the sluice network \cite{ruder2017sluice} as a special case: the sluice network predefines a fixed number of subspaces to fuse the features from different tasks between different subspaces (each contains multiple feature channels), while our model can automatically fuse the features according to \emph{each single channel}.

\subsection{Implementation Details}
Note that the state-of-the-art convolutional network architectures such as VGG \cite{simonyan2014very}, ResNet \cite{he2016deep}, and DenseNet \cite{huang2016densely} typically group similar operations into stages/blocks, where each stage contains \{convolution-activation\}$^n$ (possibly with pooling). In order to make the least modification to the baseline network architecture to investigate the performance of the proposed NDDR layer, we only connect the two networks by applying the NDDR layer \emph{at the end of each stage/block}. For example, we apply the NDDR layers at the outputs of \emph{pool1, pool2, pool3, pool4,} and \emph{pool5} for the VGG network. Similarly, as much deeper as ResNet is, we still apply only 5 NDDR layers in it, \eg, at the outputs of \emph{conv1n3, conv2\_3n3, conv3\_4n3, conv4\_6n3,} and \emph{conv5\_3n3} for ResNet-101. Also, it is worth noting that the additional parameters introduced by the NDDR layers are also very few with respect to those for the whole networks. For example, when applying NDDR layers at \emph{pool1, pool2, pool3, pool4,} and \emph{pool5} of the VGG-16 network, the additional parameters for the NDDR layers are only 1.2M, being 0.8\% compared to the original 138M parameters of the entire VGG-16.


\section{Ablation Analysis \label{Sect:AA}}
In this section, several ablations have been done to analyze NDDR-CNN. Two factors about the NDDR layers are analyzed, \ie, different $1 \times 1$ convolutional weight initializations, and the scales of the ``base'' learning rate (\ie, the learning rate for the remaining network) to train the NDDR layers. We also analyze which pretrained weights should be used as initialization, \ie, the weights trained on ImageNet or different single tasks. We use a two-task problem here and in the following sections. For the ablation analysis, we use \emph{semantic segmentation} and \emph{surface normal prediction}.

\noindent \textbf{Dataset.} The NYU v2 dataset \cite{Silberman:ECCV12} is used for semantic segmentation and surface normal prediction. We use the official train/val splits which include 795 images for training and 654 images for validation. For semantic segmentation, the NYU v2 dataset contains 40 classes such as beds, cabinets, clothes, books, \etc \cite{guptaCVPR13}. The NYU v2 dataset also has the pixel-level surface normal ground-truths precomputed by the depth labeling \cite{eigen2015predicting,Silberman:ECCV12,l2014discriminatively}. 


\noindent \textbf{Network Architecture.} We use the state-of-the-art architecture for pixel-level tasks, \ie, Deeplab \cite{chen2016deeplab}, for both semantic segmentation and surface normal prediction. Deeplab is essentially a VGG or ResNet network backbone with atrous convolution and atrous spatial pyramid pooling. We do not implement Fully Connected CRFs or multi-scale inputs as they are not related to the NDDR layer we proposed. We are careful to stick closely to the proposed NDDR layer by using the same atrous convolution and atrous spatial pyramid pooling for all the methods, so as to clearly see the effects of simply incorporating the NDDR layer. We use the Deeplab-VGG-16 architecture in all the ablation analysis. 

\noindent \textbf{Losses.} We use the softmax cross-entropy loss for semantic segmentation. For surface normal prediction, we use the $\ell_2$ regression loss after normalizing the normal vector of each pixel to have unit $\ell_2$ norm (\ie, this represents a direction for a certain angle). Therefore, our loss for surface normal prediction is also equivalent to the cosine loss.

\noindent \textbf{Evaluation Metrics.} The performance of semantic segmentation is evaluated by mean Intersection over Union (mIoU) and Pixel Accuracy (PAcc). For surface normal estimation, we use Mean and Median angle distances of all the pixels for evaluation (the lower the better). In addition, we also use the metrics introduced by \cite{eigen2015predicting}, which are the percentage of pixels that are within the angles of $11^\circ, 22.5^\circ, 30^\circ$ to the ground-truth (the higher the better).

\subsection{Initializations for NDDR Layers \label{Sect:init_fuse}}
In order to have a mild initialization which resembles single-task networks, we keep the diagonal weights of the NDDR layer as non-zeros. Recall that the NDDR layer for a two-task problem is $F^{out} =
\begin{bmatrix} F^{in}_1, F^{in}_2 \end{bmatrix} \begin{bmatrix} W_1^\top, & W_2^\top \end{bmatrix}^\top$. In order to initialize the NDDR weights $W_1$ and $W_2$, we let:
{\footnotesize
\begin{equation}
F^{out}_1 =
\begin{bmatrix}
F^{in}_1, F^{in}_2
\end{bmatrix}
\begin{bmatrix}
   \alpha & 0 & ... & 0     & \beta & 0 & ... & 0\\
   0 & \alpha & ... & 0     & 0 & \beta & ... & 0 \\
   \vdots & &\ddots &\vdots & \vdots & &\ddots &\vdots \\
   0 & 0 & ... & \alpha     & 0 & 0 & ... & \beta
\end{bmatrix}^\top,
\label{init}
\end{equation}}
where $F^{in}_1, F^{in}_2$ are the inputs to the NDDR layer and $F^{out}_1$ is the output which will be fed to Task 1\footnote{We take an NDDR layer from one task as an example, and the initialization of the NDDR layer for the other task is identical.}. By writing the weight of NDDR in this way, it shows that if we initialize $\alpha = 1$ and $\beta = 0$, the whole network will start with the single-task networks, \ie, $F^{out}_1 = F^{in}_1$. We refer to this as \emph{diagonal initialization}.

In the experiments, we have 5 different diagonal initializations with $(\alpha, \beta)$ ranging from $(1, 0)$ to $(0, 1)$, \ie, from the mildest initialization from the same tasks to the most severe initialization from the opposite tasks. In addition, we also discuss the random initialization of the whole weight matrices $[W_1^\top, W_2^\top]$ with Xavier initialization \cite{glorot2010understanding}.

Table \ref{Table:init_fuse} shows the performance with different initializations of the NDDR layer. The results show that the \emph{diagonal initialization} is better than Xavier initialization\footnote{Note that the results from Xavier initialization (in Table \ref{Table:init_fuse}) are still comparable with the previous state-of-the-art method (\ie the cross-stitch network and the sluice network in Table \ref{Table:VGG-16}) in surface normal prediction.}, and that the initialization of $(\alpha, \beta)$ has a little effect on results. In the following experiments, we use diagonal initialization with $(\alpha, \beta) = (0.9, 0.1)$.
\begin{table}[t]
\centering
\fontsize{7pt}{0.9\baselineskip}\selectfont
\begin{tabular}{@{\extracolsep{\fill}} || l || c | c | c | c | c || c | c ||}
\hline
& \multicolumn{5}{c||}{\textbf{Surface Normal Prediction}} & \multicolumn{2}{c||}{\textbf{Semantic Seg.}} \\
\hline
& \multicolumn{2}{c|}{\textbf{Angle Distance}} & \multicolumn{3}{c||}{\textbf{Within $t^\circ$ (\%)}} & \multicolumn{2}{c||}{\textbf{(\%)}} \\
& \multicolumn{2}{c|}{(Lower Better)} & \multicolumn{3}{c||}{(Higher Better)} & \multicolumn{2}{c||}{(Higher Better)} \\
\hline
($\alpha, \beta$)& Mean & Med.  & $11.25$ & $22.5$ & $30$ & mIoU & PAcc \\
\hline
(1, 0)     & 14.0 & 10.3 & 53.2 & 79.1 & 88.6 & \underline{{\bf 36.2}} & 66.5 \\
(0.9, 0.1) & \underline{{\bf 13.9}} & \underline{{\bf 10.2}}  & \underline{{\bf 53.5}} & \underline{{\bf 79.5}} & \underline{{\bf 88.8}} & \underline{{\bf 36.2}} & \underline{{\bf 66.4}} \\
(0.5, 0.5) & \underline{{\bf 13.9}} & \underline{{\bf 10.2}}  & \underline{{\bf 53.5}} & 79.3 & 88.6 & 36.0 & \underline{{\bf 66.4}} \\
(0.1, 0.9) & 14.3 & 10.6  & 52.4 & 78.5 & 88.0 & 35.7 & 66.1 \\
(0, 1)     & 14.2 & 10.6  & 52.5 & 78.2 & 87.8 & 35.7 & 65.9 \\
Random     & 15.0 & 11.6  & 49.0 & 76.7 & 87.0 & 33.4 & 64.4 \\
\hline
\end{tabular}
\caption{The results with different initializations for the \textbf{NDDR layers}.}
\label{Table:init_fuse}
\vspace{-2.5mm}
\end{table}

\begin{table}[t]
\centering
\fontsize{8pt}{0.9\baselineskip}\selectfont
\begin{tabular}{@{\extracolsep{\fill}} || l || c | c | c | c | c || c | c ||}
\hline
& \multicolumn{5}{c||}{\textbf{Surface Normal Prediction}} & \multicolumn{2}{c||}{\textbf{Semantic Seg.}} \\
\hline
& \multicolumn{2}{c|}{\textbf{Errors}} & \multicolumn{3}{c||}{\textbf{Within $t^\circ$ (\%)}} & \multicolumn{2}{c||}{\textbf{(\%)}} \\
& \multicolumn{2}{c|}{(Lower Better)} & \multicolumn{3}{c||}{(Higher Better)} & \multicolumn{2}{c||}{(Higher Better)} \\
\hline
Scale & Mean & Med. & $11.25$ & $22.5$ & $30$ & mIoU & PAcc \\
\hline
$1$     & 14.7 & 11.2 & 50.1 & 77.3 & 87.4 & 35.9 & 65.9 \\
$10$    & 14.4 & 10.7 & 51.9 & 78.1 & 87.9 & 36.0 & 66.1 \\
$10^2$  & \underline{{\bf 13.9}} & \underline{{\bf 10.2}}  & \underline{{\bf 53.5}} & 79.5 & 88.8 & \underline{{\bf 36.2}} & \underline{{\bf 66.4}} \\
$10^3$  & \underline{{\bf 13.9}} & 10.6 & 52.4 & \underline{{\bf 79.6}} & \underline{{\bf 89.2}} & 35.7 & \underline{{\bf 66.4}} \\
\hline
\end{tabular}
\caption{The results with different learning rates for the \textbf{NDDR layers} (\ie, the scale with respect to the base learning rate for other layers). The learning rates are represented as different \textbf{scales} with respect to those for other perception convolutional layers.}
\label{Table:lr_fuse}
\vspace{-3mm}
\end{table}

\subsection{Learning Rates for NDDR Layer \label{Sect:lr_fuse}}
In this section, we discuss the learning rate for the NDDR layer. There are two main reasons to set a larger learning rate specifically for the NDDR layer. First, as analyzed in Sect. \ref{Sect:init_fuse}, the NDDR-CNN becomes single-task networks if we set a very large weight (\eg, $\alpha = 1$, much larger than the weights of perception convolutional layers) at the diagonal of $W_1^\top$. Thus, we hypothesize that the magnitude of the NDDR layer weights should be larger, therefore requiring a larger learning rate. Second, a larger learning rate for NDDR layers is also necessary if we fine-tune the NDDR-CNN from the pretrained single-task networks.

Therefore, we analyze the proper learning rate for the NDDR layer as how many times it should be with respect to the base learning rate (for the remaining network excluding the NDDR layers). Table \ref{Table:lr_fuse} shows the performance of using different learning rates for the NDDR layer. It verifies that larger learning rates should be applied for NDDR layers. In the following experiments, we use 100 times of the base learning rate for the NDDR layer.
\subsection{Pretrained Weights for Network Initialization}
Two network initialization strategies can be applied. We may use the network weights pretrained on a general task (\eg pretrained Deeplab-VGG-16 \cite{chen2016deeplab} for semantic segmentation on Pascal VOC 2012 \cite{everingham2015pascal}) or finetuned on corresponding target single tasks. The results with different pretrained models are summarized in Table \ref{Table:pre_models}, which show that initializing the finetuned weights from target single tasks performs better. The results indicate that by simply adding several NDDR layers, we have enlarged the capability of the (converged) original networks, which further enables to skip the previously existing saddle points.

\begin{table}[t]
\centering
\fontsize{8pt}{0.9\baselineskip}\selectfont
\begin{tabular}{@{\extracolsep{\fill}} || l || c | c | c | c | c || c | c ||}
\hline
& \multicolumn{5}{c||}{\textbf{Surface Normal Prediction}} & \multicolumn{2}{c||}{\textbf{Semantic Seg.}} \\
\hline
& \multicolumn{2}{c|}{\textbf{Errors}} & \multicolumn{3}{c||}{\textbf{Within $t^\circ$ (\%)}} & \multicolumn{2}{c||}{\textbf{(\%)}} \\
& \multicolumn{2}{c|}{(Lower Better)} & \multicolumn{3}{c||}{(Higher Better)} & \multicolumn{2}{c||}{(Higher Better)} \\
\hline
Init. & Mean & Med. & $11.25$ & $22.5$ & $30$ & mIoU & PAcc \\
\hline
Pret.  & 14.3 & 10.6 & 52.2 & 78.6 & 88.2 & 34.3 & 65.2    \\
Sing.  & \underline{{\bf 13.9}} & \underline{{\bf 10.2}} & \underline{{\bf 53.5}} & \underline{{\bf 79.5}} & \underline{{\bf 88.8}} & \underline{{\bf 36.2}} & \underline{{\bf 66.4}}      \\
\hline
\end{tabular}
\caption{The results with different pretrained models. \textbf{Pret.} means the pretrained Deeplab-VGG-16 weights for semantic segmentation on Pascal VOC 2012, and \textbf{Sing.} represents the finetuned weights from the corresponding target single tasks (through single-task networks).}
\label{Table:pre_models}
\vspace{-5mm}
\end{table}

\vspace{-1mm}
\section{Experiments \label{Sect:exp}}
\vspace{-1mm}
In this section, we perform various experiments on both different network structures and different task sets to demonstrate the promising performance and desirable generalizability of the proposed NDDR-CNN.

Specifically, VGG-16 \cite{simonyan2014very} and ResNet-101 \cite{he2016deep} have been used in our experiments, we put the results on AlexNet \cite{krizhevsky2012imagenet} in Table \ref{Table:alexnet}.  In addition, we also test our proposed NDDR-CNN-Shortcut with the VGG structure, where the gradients can be passed to the lower NDDR layers by the shortcut connections. This can further demonstrate the performance of the proposed NDDR layer. We refer to this network as \emph{VGG-16-Shortcut}.


For evaluation, we train each task separately using the common single-task network architecture without NDDR layers as our \textbf{single task baseline}. The results from the most widely used heuristic multi-task network structure are performed as our \textbf{multi-task baseline}, where all the convolutional layers are shared and the split takes place after the last convolutional layer. We also investigate the performances of the \textbf{cross-stitch network} \cite{misra2016cross} and the state-of-the-art \textbf{sluice network} \cite{ruder2017sluice} for comparison, in which we apply the same number of cross-stitch/sluice layers at the same locations as our NDDR layers. We use the number of subspaces as 2 for sluice network as suggested in \cite{ruder2017sluice}. For the fair comparison, we use the best hyperparameters in \cite{misra2016cross} and \cite{ruder2017sluice} to train the corresponding networks\footnote{We show that the hyperparameters, originally from AlexNet in \cite{misra2016cross}, are still the best for other network backbones. Please see Table \ref{Table:vgg} in Appendix.}.

\begin{table}[t]
\centering
\fontsize{8pt}{0.9\baselineskip}\selectfont
\begin{tabular}{@{\extracolsep{\fill}} || l || c | c | c | c | c || c | c ||}
\hline
& \multicolumn{5}{c||}{\textbf{Surface Normal Prediction}} & \multicolumn{2}{c||}{\textbf{Semantic Seg.}} \\
\hline
& \multicolumn{2}{c|}{\textbf{Errors}} & \multicolumn{3}{c||}{\textbf{Within $t^\circ$ (\%)}} & \multicolumn{2}{c||}{\textbf{(\%)}} \\
& \multicolumn{2}{c|}{(Lower Better)} & \multicolumn{3}{c||}{(Higher Better)} & \multicolumn{2}{c||}{(Higher Better)} \\
\hline
& Mean & Med. & $11.25$ & $22.5$ & $30$ & mIoU & PAcc \\
\hline
Sing.  & 15.6 & 12.3 & 46.4 & 75.5 & 86.5 & 33.5 & 64.1 \\
Mul.   & 15.2 & 11.7 & 48.4 & 76.2 & 87.0 & 33.4 & 64.2 \\
C.-S.  & 15.2 & 11.7 & 48.6 & 76.0 & 86.5 & 34.8 & 65.0   \\
Sluice & 14.8 & 11.3 & 49.7 & 77.1 & 88.0 & 34.9 & 65.2   \\
Ours   & \underline{{\bf 13.9}} & \underline{{\bf 10.2}} & \underline{{\bf 53.5}} & \underline{{\bf 79.5}} & \underline{{\bf 88.8}} & \underline{{\bf 36.2}} & \underline{{\bf 66.4}}      \\
\hline
\end{tabular}
\caption{Experimental results on semantic segmentation and surface normal prediction using \textbf{VGG-16}. Sing., Mul., C.-S., and Sluice represent the single-task baseline, the multiple-task baseline, the cross-stitch network, and the sluice network, respectively.}
\label{Table:VGG-16}
\vspace{-3mm}
\end{table}

As we aim to a general purpose MTL method, very diverse task sets are chosen to evaluate our performance. These include \emph{pixel-level labeling tasks on scene images}, \ie, semantic segmentation and surface normal prediction, and \emph{image-level classification tasks on human faces}, \ie, age and gender classification. In the following subsections, we perform the semantic segmentation and surface normal prediction on NYU v2 dataset \cite{Silberman:ECCV12}, and the age and gender classification on the IMDB-WIKI dataset \cite{Rothe-IJCV-2016}. 
We detail the task configurations in the following.

\subsection{Semantic Segmentation and Surface Normal Prediction}
In this section, we test our network on VGG-16, ResNet-101, and VGG-16-Shortcut to verify the desirable performance of the proposed network. In addition, by doing this, we further demonstrate the desirable generalizability of the proposed NDDR layers on different network architectures. 

The configurations of the semantic segmentation, surface normal prediction, and the best hyperparameters to train the network can be found in Sect. \ref{Sect:AA}.
\begin{table}[t]
\fontsize{8pt}{0.9\baselineskip}\selectfont
\begin{tabular}{@{\extracolsep{\fill}} || l || c | c | c | c | c || c | c ||}
\hline
& \multicolumn{5}{c||}{\textbf{Surface Normal Prediction}} & \multicolumn{2}{c||}{\textbf{Semantic Seg.}} \\
\hline
& \multicolumn{2}{c|}{\textbf{Errors}} & \multicolumn{3}{c||}{\textbf{Within $t^\circ$ (\%)}} & \multicolumn{2}{c||}{\textbf{(\%)}} \\
& \multicolumn{2}{c|}{(Lower Better)} & \multicolumn{3}{c||}{(Higher Better)} & \multicolumn{2}{c||}{(Higher Better)} \\
\hline
& Mean & Med. & $11.25$ & $22.5$ & $30$ & mIoU & PAcc \\
\hline
Sing.   & 15.6 & 12.7 & 44.3 & 74.8 & 87.2 & 39.5 & 69.2   \\
Mult.  & 16.3 & 13.8 & 41.1 & 73.9 & 86.5 & 39.1 & 68.7   \\
C.-S.  & 15.9 & 13.2 & 42.9 & 75.1 & 86.8 & 40.5 & 69.5   \\
Sluice & 15.3 & 12.8 & 44.1 & 76.9 & 88.2 & 40.8 & 70.1   \\
Ours   & \underline{{\bf 14.4}} & \underline{{\bf 11.6}} & \underline{{\bf 48.5}} & \underline{{\bf 79.1}} & \underline{{\bf 89.5}} & \underline{{\bf 43.3}} & \underline{{\bf 71.5}}      \\
\hline
\end{tabular}
\caption{Experimental results on semantic segmentation and surface normal prediction using \textbf{ResNet-101}. Sing., Mul., C.-S., and Sluice represent the single-task baseline, the multiple-task baseline, the cross-stitch network, and the sluice network, respectively.}
\label{Table:ResNet-101}
\vspace{-3mm}
\end{table}

\subsubsection{Experiments on VGG-16 Network \label{Sect:VGG}}
In this section, we combine two VGG-16 networks by applying the NDDR layer at the outputs of \emph{pool1, pool2, pool3, pool4} and \emph{pool5}.

Table \ref{Table:VGG-16} shows the results on semantic segmentation and surface normal prediction using the VGG-16 network. Though as simple as our method is, it significantly outperforms the state-of-the-art methods. For example, our method outperforms the sluice network by around 3.8\% in ``within 11.25$^\circ$'' metric in surface normal prediction, and 1.1\%-1.2\% for both metrics in semantic segmentation. These results demonstrate the promising performance of our method.

\subsubsection{Experiments on ResNet-101 Network \label{sect:ResNet}} 
We perform the NDDR layers in the ResNet-101 network, where the NDDR layers are only applied at the output of \emph{conv1n3, conv2\_3n3, conv3\_4n3, conv4\_6n3} and \emph{conv5\_3n3}.

The results are shown in Table \ref{Table:ResNet-101}. It indicates that our method consistently outperforms the baseline and state-of-the-art results. Noted that comparing with the as deep as 101-layer network, we only slightly modified the ResNet-101 by adding \emph{five} NDDR layers. These results further demonstrate the efficacy of the proposed NDDR layer.

\begin{table}[t]
\fontsize{8pt}{0.9\baselineskip}\selectfont
\begin{tabular}{@{\extracolsep{\fill}} || l || c | c | c | c | c || c | c ||}
\hline
& \multicolumn{5}{c||}{\textbf{Surface Normal Prediction}} & \multicolumn{2}{c||}{\textbf{Semantic Seg.}} \\
\hline
& \multicolumn{2}{c|}{\textbf{Errors}} & \multicolumn{3}{c||}{\textbf{Within $t^\circ$ (\%)}} & \multicolumn{2}{c||}{\textbf{(\%)}} \\
& \multicolumn{2}{c|}{(Lower Better)} & \multicolumn{3}{c||}{(Higher Better)} & \multicolumn{2}{c||}{(Higher Better)} \\
\hline
& Mean & Med. & $11.25$ & $22.5$ & $30$ & mIoU & PAcc \\
\hline
Sing.  & 15.5 & 12.3 & 46.3 & 75.5 & 86.5 & 33.5 & 64.4 \\
Mult.  & 15.2 & 11.8 & 48.3 & 76.1 & 86.6 & 33.6 & 64.4 \\
C.-S.  & 14.8 & 11.1 & 50.3 & 76.9 & 87.0 & 35.0 & 65.1   \\
Sluice & 14.2 & 10.6 & 51.7 & 78.2 & 88.2 & 35.3 & 65.3   \\
Ours   & \underline{{\bf 13.5}} & \underline{{\bf 9.8}} & \underline{{\bf 55.3}} & \underline{{\bf 80.5}} & \underline{{\bf 89.3}} & \underline{{\bf 36.7}} & \underline{{\bf 67.0}}      \\
\hline
\end{tabular}
\caption{Experimental results on semantic segmentation and surface normal prediction using \textbf{VGG-16-Shortcut}. Sing., Mul., C.-S., and Sluice represent the single-task baseline, the multiple-task baseline, the cross-stitch network, and the sluice network, respectively.}
\label{Table:VGG-16-Shortcut}
\vspace{-3mm}
\end{table}

\subsubsection{Experiments on VGG-16 Network with Shortcut Connections}
We test the proposed NDDR-CNN-Shortcut with the VGG-16 structure, \ie, the VGG-16-Shortcut network to further validate our performance.

Compared with ResNet, the VGG-16-Shortcut network resembles more to DenseNet \cite{huang2016densely}. In VGG-16-Shortcut, the gradients can be passed to the lower NDDR layers by the \emph{direct and shortest} shortcut connections, rather than by \emph{multiple} shortcuts in ResNet-like networks where the gradients may still decay\footnote{Note that we did not implement the ResNet-like shortcuts, such as DenseNet. This is because that the ResNet-like shortcuts in DenseNet is less related to the NDDR layer we proposed. Therefore, we carefully stitch to the factors that influence the NDDR layer to analysis it more clearly.}.

The results for VGG-16-Shortcut are shown in Table \ref{Table:VGG-16-Shortcut}. Compared with the performance on the ``vanilla'' VGG-16 network (\ie, Table \ref{Table:VGG-16}), the results of all the methods are improved in VGG-16-Shortcut. Especially, the improvements in our method are higher than those in our counterpart. 

Table \ref{Table:VGG-16-Shortcut} shows that our method consistently outperforms the state-of-the-art methods. Especially, our method outperforms the sluice network by 3.1\% for ``within 11.25$^\circ$'' metric in surface normal prediction, and 1.0\%-1.5\% for the two metrics in semantic segmentation.

\begin{figure}[t]
\centering
\includegraphics[width=\linewidth]{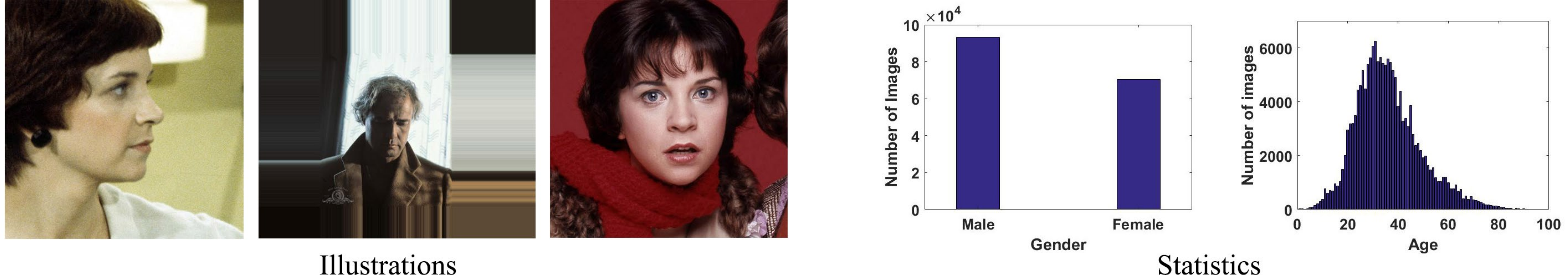}
\caption{Some example illustrations and statistics of ages and genders for the IMDB-WIKI dataset. The statistics show that we have sufficient samples to train both genders, and the ages of most samples are between 20 - 50.}
\label{Fig:IMDB}
\vspace{-3mm}
\end{figure}

\subsection{Age Estimation and Gender Classification} 
\noindent \textbf{Dataset.} We use the IMDB-WIKI dataset \cite{Rothe-IJCV-2016} for this task set, which contains 460723 images collected from 20284 subjects. After filtering out the images with more than one faces and the images without age or gender labels, the remaining 187103 images from 12325 subjects are used to perform our experiments. These contain images for both genders with ages from 0 to 99. We randomly choose 24090 images from 2000 subjects for evaluation, and the remaining 163013 images from 10325 subjects are used for training. In the training set, we have sufficient samples for both male and female, but the training data for ages is imbalanced. Some image examples, with the gender and age statistics, are shown in Fig. \ref{Fig:IMDB}.

\noindent \textbf{Network Architecture.} The VGG-16 network is used as the base network in this experiment, with the NDDR layers applied after \emph{pool1, pool2, pool3, pool4} and \emph{pool5}.

\noindent \textbf{Losses.} Motivated by \cite{Rothe-IJCV-2016}, we treat both age and gender estimations as classification problems, \ie, 2-class and 100-class classifications. We use softmax cross-entropy loss in both tasks.

\noindent \textbf{Evaluation Metrics.} Classification accuracy (Acc) is used to evaluate the gender classification. For age estimation, we follow the metric from \cite{Rothe-IJCV-2016}. That is, for each image $i$, we treat the output $p_i \in \mathbb{R}^{100}$ from softmax as the probabilities for different ages (\ie, 0-99). Therefore the final age estimation is calculated by $\text{age}_i^* = \sum_{k=0}^{99} p_i(k) \text{dict}(k)$, where dict = $\{0, 1, ..., 99 \} \in \mathbb{R}^{100}$ is the age dictionary. We use Mean Absolute Error (Mean AE) and Median Absolute Error (Median AE) for evaluating the age estimation.

The experimental results are show in Table \ref{Table:age_gender}. It shows that our method on age estimation significantly outperforms the state-of-the-art methods, \ie, $(8.5-8.0)/8.5 \approx 5.9\%$ for Mean AE and $(7.0-6.2)/7.0 \approx 11.4\%$ for Median AE. While for the gender classification, our method just performs comparably with the cross-stitch network. This is because that gender classification is a two-class classification problem with sufficient labeled samples for each gender. Therefore, it benefits less from the other task (with another set of labels) when learning the representation.
\begin{table}[t]
\centering
\scalebox{0.8}{
\begin{tabular}{@{\extracolsep{\fill}} || l || c | c || c ||}
\hline
& \multicolumn{2}{c||}{\textbf{Age}} & \textbf{Gender} \\
& \multicolumn{2}{c||}{(Lower Better)} & \multicolumn{1}{c||}{(Higher Better)} \\
\hline
& Mean AE & Median AE & Acc. (\%)  \\
\hline
Single-Task                  & 9.1 & 7.4 & 83.5    \\
Multi-Task                   & 9.0 & 7.4 & 82.3    \\
Cross-Stitch                 & 8.6 & 7.0 & \underline{{\bf 84.0}}    \\
Sluice                       & 8.5 & 7.0 & 83.9    \\
Ours                         & \underline{{\bf 8.0}} & \underline{{\bf 6.2}} & \underline{{\bf 84.0}}    \\
\hline
\end{tabular}}
\caption{Experimental results on age and gender classification.}
\label{Table:age_gender}
\vspace{-3mm}
\end{table}

\section{Conclusions \label{Sect:conclusion}}
In this paper, we proposed a novel CNN structure for general-purpose MTL. Firstly, the task-specific features with the same spatial resolution from different tasks were concatenated. Then, we performed Neural Discriminative Dimensionality Reduction (NDDR) over them to learn a discriminative feature embedding for each task, which also satisfies input sizes of the following layers. 

The NDDR layer is simple and effective, which is constructed by combining existing CNN components in a novel way. The proposed NDDR networks can be trained in an end-to-end fashion without any extraordinary operations of a modern CNN. This desirable property guarantees that the proposed NDDR layer can easily be extended to various state-of-the-art CNN architectures in a ``plug-and-play'' manner. In addition, our proposed NDDR-CNN generalizes several state-of-the-art CNN based MTL models, such as the cross-stitch network \cite{misra2016cross} and the sluice network \cite{ruder2017sluice}.

We performed detailed ablation analysis, showing that the proposed NDDR layer is easy to train and also robust to different hyperparameters. The experiments on various CNN structures and different task sets demonstrate the promising performance and desirable generalizability of our proposed method. An interesting future research direction can be studying explicitly imposing various dimensionality reduction assumptions on the NDDR layer.

\renewcommand\thesubsection{A\arabic{subsection}}
\renewcommand{\thefigure}{A\arabic{figure}}
\setcounter{figure}{0}
\renewcommand{\thetable}{A\arabic{table}}
\setcounter{table}{0}
\section*{Appendix \label{sect:app}}
We conduct additional experiments in this section including:
\begin{itemize}
    \item Semantic Segmentation and Surface Normal Prediction using AlexNet \cite{krizhevsky2012imagenet} backbone.
    \item Additional ablation analysis for the cross-stitch network on VGG-16 backbone \cite{simonyan2014very} to verify the hyperparameters for the cross-stitch network used in our main text are optimal.
\end{itemize}

\begin{table}[!htp]
\centering
\fontsize{7pt}{0.9\baselineskip}\selectfont
\begin{tabular}{@{\extracolsep{\fill}} || l || c | c | c | c | c || c | c ||}
\hline
& \multicolumn{5}{c||}{\textbf{Surface Normal Prediction}} & \multicolumn{2}{c||}{\textbf{Semantic Seg.}} \\
\hline
& \multicolumn{2}{c|}{\textbf{Angle Dist.}} & \multicolumn{3}{c||}{\textbf{Within $t^\circ$ (\%)}} & \multicolumn{2}{c||}{\textbf{(\%)}} \\
\hline
\textbf{AlexNet} & Mean & Med.  & $11.25$ & $22.5$ & $30$ & mIoU & PAcc \\
\hline
C.-S.   & 19.7 & 17.1 & 28.1 & 65.9 & \underline{{\bf 80.0}} & 21.7 & 53.4 \\
Sluice  & 19.5 & 16.6 & 29.7 & 66.2 & 79.5 & 21.9 & 53.8 \\
Ours       & \underline{{\bf 19.4}} & \underline{{\bf 15.5}}  & \underline{{\bf 36.6}} & \underline{{\bf 66.8}} & 79.2 & \underline{{\bf 23.1}} & \underline{{\bf 56.3}} \\
\hline
\end{tabular}
\caption{The results for \emph{Semantic Segmentation} and \emph{Surface Normal Prediction} on \textbf{AlexNet}.}
\label{Table:alexnet}
\vspace{-3mm}
\end{table}

\begin{table}[!tph]
\centering
\fontsize{7pt}{0.9\baselineskip}\selectfont
\begin{tabular}{@{\extracolsep{\fill}} || l || c | c | c | c | c || c | c ||}
\hline
& \multicolumn{5}{c||}{\textbf{Surface Normal Prediction}} & \multicolumn{2}{c||}{\textbf{Semantic Seg.}} \\
\hline
& \multicolumn{2}{c|}{\textbf{Angle Dist.}} & \multicolumn{3}{c||}{\textbf{Within $t^\circ$ (\%)}} & \multicolumn{2}{c||}{\textbf{(\%)}} \\
\hline
\hline
($\alpha, \beta$)& Mean & Med.  & $11.25$ & $22.5$ & $30$ & mIoU & PAcc \\
\hline
(0.9, 0.1) & \underline{{\bf 15.2}} & 11.7  & 48.6 & \underline{{\bf 76.0}} & \underline{{\bf 86.5}} & \underline{{\bf 34.8}} & \underline{{\bf 65.0}} \\
(0.7, 0.3) & 15.5 & \underline{{\bf 11.6}} & \underline{{\bf 48.7}} & 75.1 & 85.5 & 34.4 & 64.6 \\
(0.5, 0.5) & 15.9 & 12.0 & 47.5 & 73.7 & 84.4 & 33.9 & 64.0 \\
\hline
\hline
Scale & Mean & Med. & $11.25$ & $22.5$ & $30$ & mIoU & PAcc \\
\hline
1          & 15.3 & 11.9  & 47.9 & 75.8 & 86.3 & 34.5 & 64.6 \\
10         & 15.5 & 12.0  & 47.3 & 75.1 & 86.0 & 35.0 & 65.0 \\
10$^2$     & 15.3 & 11.8  & 48.1 & 75.6 & 86.2 & \underline{{\bf 35.1}} & \underline{{\bf 65.2}} \\
10$^3$     & \underline{{\bf 15.2}} & \underline{{\bf 11.7}}  & \underline{{\bf 48.6}} & \underline{{\bf 76.0}} & \underline{{\bf 86.5}} & 34.9 & 65.0 \\
\hline
\end{tabular}
\caption{Ablation analysis for the cross-stitch network on \textbf{VGG-16}. This is to ensure that the hyperparameters for the cross-stitch network, \ie, $(\alpha, \beta) = (0.9, 0.1)$ and 1000x learning rate for fuse layers, used in our main text are the best ones for the cross-stitch network.}
\label{Table:vgg}
\end{table}

\subsection{Semantic Segmentation and Surface Normal Prediction on AlexNet}
We conduct \emph{Semantic Segmentation} and \emph{Surface Normal Prediction} on AlexNet \cite{krizhevsky2012imagenet} with FCN32s \cite{long2015fully}, as those in the cross-stitch network paper \cite{misra2016cross}. We also use the same hyperparameters as the those in \cite{misra2016cross}. The results in Table \ref{Table:alexnet} show that our method outperforms the cross-stitch network and the sluice network on AlexNet.

\subsection{Ablation Analysis for the Cross-Stitch Network on VGG-16}
In this section, we verify that, in our main text, we have fair comparisons with the state-of-the-art cross-stitch network, especially regarding the hyperparameters on different network backbones. In other words, we show that the hyperparameters for the cross-stitch network, originally obtained from \cite{misra2016cross} on AlexNet, are still the best for other network backbones. This can be investigated by doing ablation analysis of the cross-stitch network on other network backbones. The ablation analysis of the cross-stitch network on VGG-16 \cite{simonyan2014very} is shown in Table \ref{Table:vgg}, which demonstrates that the best hyperparameters of the cross-stitch network have been used in our main text for fair comparative-evaluation.

\section*{Acknowledgements}
This work is partially supported by NSFC 61773295, NSFC 61601112, ONR N00014-12-1-0883.

{
\bibliographystyle{ieee_fullname}
\bibliography{egbib2}
}

\end{document}